\documentclass[11pt]{article}

\usepackage{acl}

\usepackage{times}
\usepackage{latexsym}
\usepackage[T1]{fontenc}
\usepackage[utf8]{inputenc}
\usepackage{microtype}
\usepackage{inconsolata}
\usepackage{graphicx}
\usepackage{xspace}
\usepackage{amsmath,amssymb}
\usepackage{booktabs}
\usepackage{multirow}
\usepackage{tabularx}
\usepackage{enumitem}
\usepackage{xcolor}
\usepackage{colortbl}
\usepackage{algorithm}
\usepackage{algpseudocode}
\usepackage{pifont}
\usepackage{listings}
\usepackage[most]{tcolorbox}
\usepackage{placeins}

\newcommand{\method}{\textsc{GAPD}\xspace}
\definecolor{gapdrow}{HTML}{FFF2CD}
\newcommand{\methodfull}{\textbf{G}old-\textbf{A}ction \textbf{P}olicy \textbf{D}istillation\xspace}
\newcommand{\kbqarone}{\textsc{KBQA-R1}\xspace}
\newcommand{\kbqaoone}{\textsc{KBQA-O1}\xspace}

\newcommand{\rrs}{\textsc{RRS}\xspace}
\newcommand{\grpo}{\textsc{GRPO}\xspace}
\newcommand{\midanchor}{\textsc{Entity-Anchor Matching}\xspace}

\newcommand{\smallpm}[1]{{\fontsize{5pt}{5pt}\selectfont $\pm$#1}}

\title{\method: \methodfull for Agentic Reinforcement Learning in Knowledge Base Question Answering}

\author{
\textbf{Xin Sun$^{1,2}$ \quad
Jianan Xie$^{3}$ \quad
Zhongqi Chen$^{4}$ \quad
Qiang Liu$^{2}$} \\
\textbf{Shu Wu$^{2}$ \quad
Bowen Song$^{4}$ \quad
Weiqiang Wang$^{4}$ \quad
Zilei Wang$^{1}$ \quad
Liang Wang$^{2}$} \\
$^{1}$University of Science and Technology of China \\
$^{2}$NLPR, MAIS, Institute of Automation, Chinese Academy of Sciences \\
$^{3}$ShanghaiTech University \quad
$^{4}$Ant Group \\
\texttt{xin.sun@cripac.ia.ac.cn}, \texttt{xiejn2025@shanghaitech.edu.cn} \\
\texttt{\{qiang.liu, shu.wu, wangliang\}@nlpr.ia.ac.cn}, \texttt{zlwang@ustc.edu.cn} \\
\texttt{\{chenzhongqi.czq, bowen.sbw, weiqiang.wwq\}@antgroup.com}
}

\begin{document}
\maketitle

\begin{abstract}
Reinforcement learning (RL) is a natural fit for agentic knowledge base question answering (KBQA), where a model must issue executable actions, observe knowledge-base feedback, and eventually return an answer.
However, current RL-based KBQA systems mainly optimize sparse rewards from the final answer, leaving intermediate action errors weakly supervised.
This is especially limiting for logical-form annotated KBQA benchmarks: gold logical forms can be converted into executable action sequences, but existing pipelines use them mainly for warm-start data construction rather than for on-policy RL updates.
We propose \textbf{\method}, a training-time \methodfull framework that adds dense token-level guidance to outcome-based RL.
To align gold actions with on-policy student rollouts, \method uses \midanchor: it treats the intermediate entities reached during student exploration and gold execution as state anchors, and matches student states to gold states through these explored entity sets.
The current policy conditioned on this aligned gold action serves as a stop-gradient teacher, whose token distribution is distilled back to the ordinary student policy over matched student-generated token spans.
\method consistently surpasses the current state of the art on WebQSP, GrailQA, and GraphQ.
\end{abstract}

\section{Introduction}

Knowledge Base Question Answering (KBQA) requires a model to answer natural language questions by navigating a structured knowledge base (KB), such as Freebase~\cite{Freebase}, and producing executable logical operations over entities, relations, and constraints~\cite{SP,WebQSP,GrailQA}.

From the perspective of external-knowledge agents, KG agents for KBQA are the structured counterpart of retrieval-augmented generation and search agents.
RAG-style systems retrieve passages or documents from unstructured corpora, using dense retrieval and retrieval-augmented language modeling or generation as core mechanisms~\cite{karpukhin2020dense,guu2020realm,RAG,izacard2021fid,borgeaud2022retro}.
Search-agent variants similarly retrieve web evidence for reasoning~\cite{xiong2025rag}, while recent work studies reliability and multimodal evidence in retrieved contexts~\cite{sun-etal-2025-divide,sun2026predict,du2026multimodal}.
KG agents instead query a structured graph: each action selects an entity, relation, operator, or constraint and changes the executable state.
This makes relation quality and precise graph navigation central, since a wrong relation or constraint can propagate through later execution; robust relation extraction is one source of reliable relational facts for such KBs~\cite{sun-etal-2023-noise}.

Prior KBQA work spans semantic parsing~\cite{SP}, subgraph-based reasoning~\cite{NSM}, and LLM-based logical-form generation~\cite{ChatKBQA}.
Recent agentic formulations instead decompose KBQA into interactions with an executable environment~\cite{Interactive-KBQA,luo2025kbqa,knowcoder-a1, kbqa-r1}.
In this setting, the model emits an action, the KB executor returns an observation, and the model continues until it can produce an answer.
Compared with one-shot logical-form generation, this action-based formulation can improve robustness because intermediate actions can be executed and checked against the KB environment.
This setting is also a specialized form of tool-augmented generation: the model does not choose among open-ended APIs, but selects KB operations from a fixed action space.
Prior tool-learning work studies scalable tool representation and selection as tool libraries grow~\cite{su-etal-2025-toolscaler,fang2026toolweaver}; our focus is how to train structured KB actions with token-level RL supervision.

Reinforcement learning (RL) is a natural fit for such agentic KBQA, and has recently become a central post-training method for improving LLM reasoning with verifiable rewards~\cite{schulman2017proximal,shao2024deepseekmath,guo2025deepseek,yu2025dapo}.
Systems such as \kbqarone~\cite{kbqa-r1} and KnowCoder-A1~\cite{knowcoder-a1} train policies with outcome-based rewards, typically using answer F1 and format validity, and update the policy with algorithms such as \grpo~\cite{shao2024deepseekmath}.
This shifts learning from static trace imitation to verifiable interaction optimization.
\textbf{However, in outcome-reward GRPO, supervision is attached to the whole rollout: generated tokens in the same response inherit the same group-relative advantage computed from the final answer reward.}
When a trajectory fails, this rollout-level signal gives little information about which intermediate action caused the failure or which token should have been changed.

Logical-form annotated KBQA benchmarks already provide executable programs or semantic parses, from which gold action sequences can be derived~\cite{SP,WebQSP,GrailQA,GraphQ}.
Prior KBQA pipelines have used this information for few-shot logical-form demonstrations~\cite{KB-BINDER,KB-Coder}, offline program or logical-form supervision~\cite{ProgramTransfer,TIARA,ChatKBQA}, trajectory synthesis~\cite{luo2025kbqa}, or supervised warm-starting~\cite{kbqa-r1}.
However, prior RL-based KBQA methods typically discard this supervision during on-policy updates and optimize only final outcome reward~\cite{kbqa-r1,knowcoder-a1}.

We address this gap by introducing \textbf{\method} (\methodfull), a framework for RL-based KBQA.
The core idea is simple: when the student policy reaches an intermediate state, we first decide which gold action is valid for that state, use the aligned action to condition the current model as a teacher, and distill the teacher's token-level preference back into the student policy.
This turns gold actions into dense guidance without replacing RL exploration with direct behavior cloning.
The teacher is not a separate frozen oracle; it is the current policy conditioned on additional gold-action information.
Thus, the distillation signal asks the model to move from its unguided behavior toward its own behavior under valid procedural conditioning.
The alignment step is necessary because on-policy rollouts can deviate from the gold trajectory: the next gold action should be determined by the current execution state, not by the raw turn index.
\method therefore uses \midanchor inside the distillation procedure.
It represents each student and gold intermediate state by the entities reached during execution and matches the student state to the most compatible gold state.

Our contributions are:
\begin{itemize}[leftmargin=*]
    \item[\ding{182}] We identify the token-level credit-assignment problem in agentic KBQA RL policies.
    \item[\ding{183}] We propose \method, which uses \midanchor to align gold actions with  student states, constructs gold-action-conditioned self-teacher distributions, and supplies token-level distillation guidance during RL.
    \item[\ding{184}] We show that \method surpasses the current state of the art on WebQSP, GrailQA, and GraphQ.
\end{itemize}

\section{Background}

\subsection{Agentic KBQA}

We follow the action-based, agentic KBQA setting explored by \kbqaoone, \kbqarone, and KnowCoder-A1~\cite{luo2025kbqa,kbqa-r1,knowcoder-a1}.
Given a question $q$, linked topic entities, and a KB executor, the model generates a sequence of actions such as \texttt{Find\_relation}, \texttt{Merge}, \texttt{Order}, \texttt{Compare}, \texttt{Time\_constraint}, and \texttt{Count}.
Appendix~\ref{sec:appendix_action_space} provides the full action templates and their corresponding logical-form updates.
Each action updates an intermediate S-expression state.
To obtain KB feedback, the executor translates the current S-expression into SPARQL and executes the SPARQL query against the KB.
The environment returns observations containing retrieved entities, relation candidates, execution status, and diagnostic messages.
The trajectory terminates when the model emits an answer.

\subsection{Outcome-Only RL}

Let $\pi_\theta$ be the student policy and $\tau = (a_1,o_1,\ldots,a_T,o_T,\hat{y})$ be a rollout.
Prior RL-based KBQA training assigns a trajectory-level reward:
\begin{equation}
R(\tau) =
\lambda_{\mathrm{out}} r_{\mathrm{out}}(\hat{y}, y^*)
+ \lambda_{\mathrm{fmt}} r_{\mathrm{fmt}}(\tau),
\end{equation}
where $r_{\mathrm{out}}$ measures final answer correctness and $r_{\mathrm{fmt}}$ rewards valid output format.
\grpo estimates the advantage by comparing multiple sampled rollouts for the same prompt:
\begin{equation}
\hat{A}_i = R(\tau_i) - \frac{1}{n}\sum_{j=1}^{n}R(\tau_j).
\end{equation}
Here $i$ indexes a sampled multi-turn response, where a response denotes the whole rollout rather than a single KBQA turn.
This objective can improve final answers, but its credit assignment is coarse: all generated tokens are weighted by the same trajectory advantage.

\subsection{Gold Actions in Existing Pipelines}

Logical-form annotated KBQA datasets provide a gold logical form $z^*$.
An action parser can convert $z^*$ into a gold action sequence
$\mathbf{g} = (g_1,\ldots,g_M)$.
Prior KBQA pipelines use this supervision in several offline ways.
Supervised semantic parsers and LLM-based generators learn from question--program or question--logical-form pairs~\cite{ProgramTransfer,TIARA,RnG-KBQA,GMT-KBQA,FC-KBQA,ChatKBQA}.
Few-shot prompting methods instead use labeled logical forms as demonstrations without fine-tuning~\cite{KB-BINDER,KB-Coder}.
Agentic methods use stepwise traces more directly: \kbqaoone initializes policy and reward models from annotated trajectories and further fine-tunes them with MCTS-generated annotations~\cite{luo2025kbqa}, while \kbqarone uses referenced rejection sampling (\rrs), where gold actions extracted from S-expressions guide trajectory synthesis before SFT~\cite{kbqa-r1}.
KnowCoder-A1 instead emphasizes outcome supervision and curriculum RL to avoid process-supervised trajectory imitation~\cite{knowcoder-a1}.
These strategies improve prompting, warm-start quality, or outcome-supervised exploration, but gold actions do not directly affect the later RL policy gradient.
Our work uses gold actions during RL itself.

\section{Method}

\subsection{Overview}

\method augments outcome-based RL for KBQA~\cite{kbqa-r1,knowcoder-a1} with dense supervision derived from gold action sequences.
For each training question, the dataset provides a gold logical form or its corresponding executable action list, as in standard KBQA semantic-parsing benchmarks~\cite{SP,WebQSP,GrailQA,GraphQ}.
During RL, the student policy still performs on-policy KB interaction and receives the original outcome reward.
In addition, for matched intermediate states, \method constructs a teacher-side prompt by conditioning the current policy on the aligned gold action, obtains a self-teacher distribution, and distills this distribution back to the student tokens.

\begin{figure*}[t]
    \centering
    \includegraphics[width=0.97\textwidth,keepaspectratio]{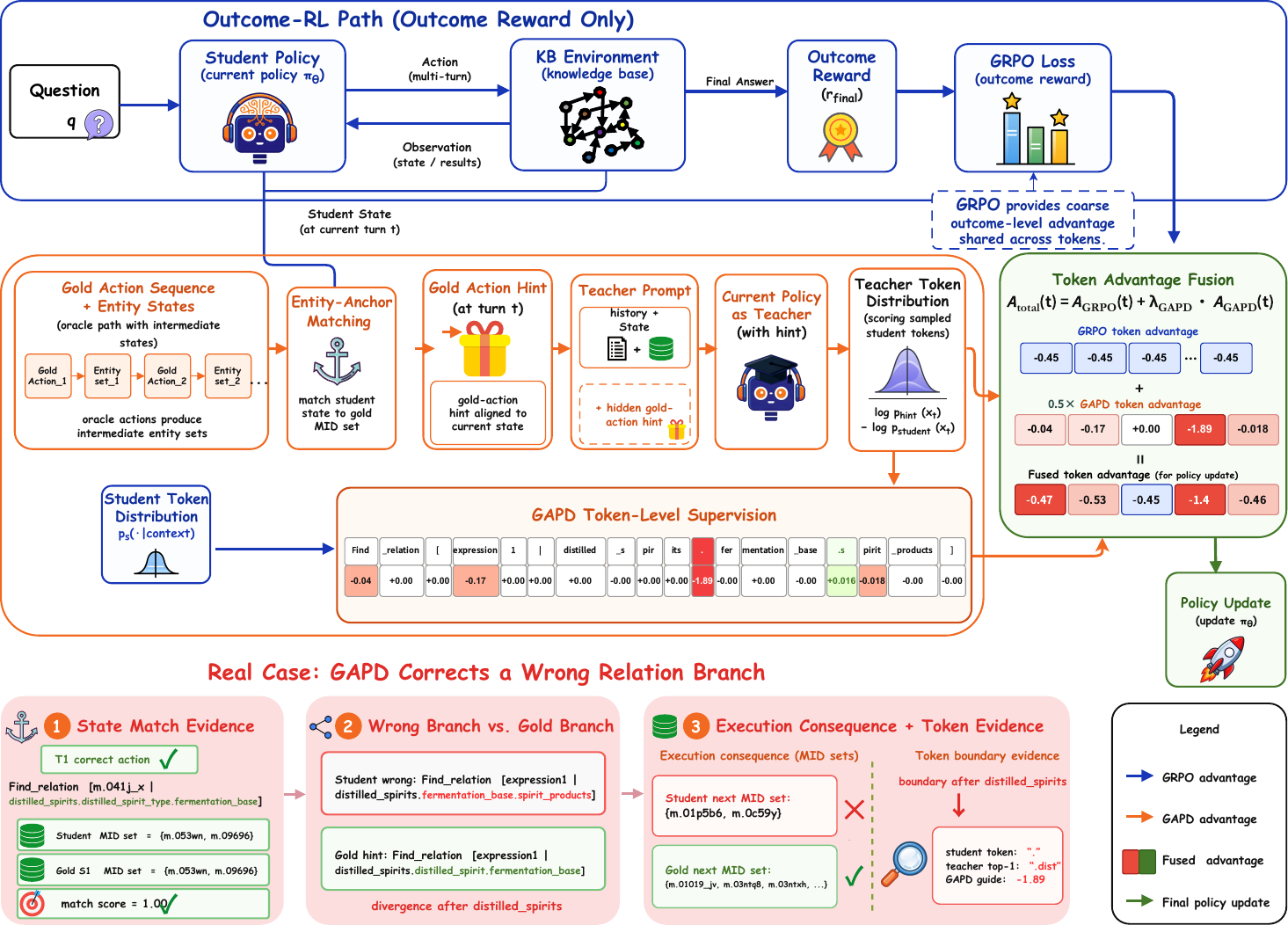}
    \caption{Overview of \method. GRPO supplies supervision mainly through the final answer reward. \method uses \midanchor to align an on-policy student state with a gold execution state, conditions the current policy on the aligned gold action to obtain a self-teacher distribution, assigns dense guidance to the selected student tokens, and fuses this guidance with the original GRPO advantage.}
    \label{fig:gapd_framework}
\end{figure*}

Figure~\ref{fig:gapd_framework} contrasts the two supervision signals used in training.
The blue path shows the original outcome-RL signal, where the final answer reward is broadcast as a coarse GRPO advantage.
The orange and green paths show \method: \midanchor first aligns a student state with a gold execution state, the aligned gold action conditions the teacher-side context, the gold-action-conditioned self-teacher scores the sampled student tokens, and the resulting token-level guide is fused with the GRPO advantage before the policy update.
The detailed training procedure is provided in Appendix~\ref{sec:appendix_training_algorithm} as Algorithm~\ref{alg:gapd}.

\subsection{Gold-Action-Conditioned Self-Teacher}
\label{sec:gapd_teacher}

As in prior action-based KBQA pipelines~\cite{luo2025kbqa,kbqa-r1}, we parse the gold logical form $z^*$ to obtain a ground-truth action sequence $\mathbf{g}=(g_1,\ldots,g_M)$, where each $g_m$ is an atomic operation in the action space.
We execute this sequence once to collect the corresponding gold execution states $u_m$ and their reached entity sets $E(u_m)$.
These gold states are used only for training-time alignment and supervision construction; they are not exposed to the student at inference time.
The orange self-teacher path in Figure~\ref{fig:gapd_framework} visualizes this construction: a gold action sequence is paired with intermediate entity anchors, \midanchor selects the aligned state, and the next gold action is inserted only into the teacher-side prompt for scoring.

At a student turn $t$, let $x_t$ be the ordinary student context containing the original question, previous student actions, and previous KB observations.
Let $m^*(t)$ denote the accepted gold-state index selected by \midanchor for the student state $s_t$.
When this match is accepted, the next gold action $g_{m^*(t)+1}$ provides a candidate procedural conditioning signal.
We construct a teacher-side prompt $\tilde{x}_t$ by adding this gold action to the same student context:
\begin{equation}
\tilde{x}_t = \mathrm{GoldCond}(x_t, g_{m^*(t)+1}).
\end{equation}
The added action describes the next valid operation in the action language.
It is intentionally procedural: it may reveal the next relation, merge target, or constraint operation, but it does not directly reveal the final answer.

The self-teacher is the current policy conditioned on the gold-action prompt:
\begin{equation}
p_{\theta}^{\mathrm{teach}}(\cdot \mid \tilde{x}_t)
= \mathrm{stopgrad}\left(\pi_\theta(\cdot \mid \tilde{x}_t)\right).
\end{equation}
The student distribution is the same policy conditioned on the original prompt:
\begin{equation}
p_{\theta}^{\mathrm{stud}}(\cdot \mid x_t)
= \pi_\theta(\cdot \mid x_t).
\end{equation}
This design follows the on-policy distillation view that the student should receive feedback on tokens sampled from its own current rollout~\cite{agarwal2024gkd,gu2024minillm,song2026surveyopd}.
Unlike hard behavior cloning, the target is not a single gold string but the model's own distribution under valid procedural conditioning.
The student is therefore encouraged to move toward a gold-action-conditioned self-teacher while still being optimized by outcome reward, preserving softer preferences among generated-token alternatives.

\subsection{Gold-Action Selection with \midanchor}
\label{sec:midanchor}

Gold actions are useful for distillation only when they correspond to the student's current execution state.
In an executable KBQA environment, the state after each action determines which next operations are valid: a relation, merge target, or constraint that is correct for one intermediate entity set may be invalid for another.
Thus, \method must align gold actions by state rather than blindly applying the gold action at the same turn index.
\midanchor performs this alignment using the intermediate entity sets reached during student exploration and gold execution.
The state-alignment box and the lower real-case panel in Figure~\ref{fig:gapd_framework} show this principle: when the student and gold entity anchors match, the next gold action is treated as the action that should be valid from that state; when execution later diverges, the teacher can penalize the specific relation branch that led to the wrong next entity set.
For a student state $s_t$ and a gold state $u_m$, \midanchor computes the Jaccard similarity over their reached entity sets:
\begin{equation}
\mathrm{sim}(s_t,u_m)=
\frac{|E(s_t)\cap E(u_m)|}{|E(s_t)\cup E(u_m)|}.
\end{equation}
The matched gold state is selected as
\begin{equation}
m^*(t)=\arg\max_m \mathrm{sim}(s_t,u_m).
\end{equation}
If this match is accepted, the next gold action $g_{m^*(t)+1}$ becomes the conditioning action for the teacher-side prompt.
\midanchor is part of training-time teacher construction, not an additional reasoning module at inference time.
We apply two conservative filters before assigning token-level guidance.
First, we skip answer turns because the teacher conditioning uses a next reasoning action, while an answer turn is already supervised by the final outcome reward.
Second, we use a same-next-state filter: if the executed student action reaches the same non-empty next entity set as the aligned gold action, we drop the distillation signal for that turn.
This preserves exploration of equivalent KB paths instead of forcing every successful state transition onto the annotated gold surface path.

\subsection{Token-Level Fusion with GRPO}
\label{sec:gapd_fusion}

\method does not optimize a standalone JSD regularizer.
Instead, it converts the gold-action-conditioned self-teacher distribution into a dense token-level guide and fuses this guide with the original GRPO policy update~\cite{shao2024deepseekmath}.
Let $y_{i,k}$ be the sampled token at position $k$ of response $i$, and let $\mathcal{I}$ denote the set of selected student-generated token positions after \midanchor matching.
In Figure~\ref{fig:gapd_framework}, this is the transition from the orange ``GAPD Token-Level Supervision'' strip to the green ``Token Advantage Fusion'' box: GAPD produces token-specific scores only on matched generated spans, while the blue GRPO advantage remains the outcome-level signal shared across response tokens.
For $k\in\mathcal{I}$, we score the sampled token under the gold-action-conditioned self-teacher and the ordinary student context:
\begin{align}
d_{i,k}
&= \log p_{\theta}^{\mathrm{teach}}
(y_{i,k}\mid \tilde{x}_{i,k}) \nonumber\\
&\quad - \log p_{\theta}^{\mathrm{stud}}
(y_{i,k}\mid x_{i,k}).
\end{align}
We use this score as a detached, clipped token advantage:
\begin{equation}
\label{eq:gapd_advantage}
A_{i,k}^{\mathrm{gapd}}
= \mathrm{clip}\left(\mathrm{sg}(d_{i,k}), -c, c\right),
\end{equation}
and $A_{i,k}^{\mathrm{gapd}}=0$ for tokens outside $\mathcal{I}$.
A positive value reinforces a sampled token that the gold-action-conditioned self-teacher assigns higher probability than the ordinary student context, while a negative value suppresses a token that the self-teacher disfavors.

Let $A_i^{\mathrm{grpo}}$ be the standard group-relative advantage derived from the final answer reward for response $i$.
We form a token-level fused advantage
\begin{equation}
\widehat{A}_{i,k}
= A_i^{\mathrm{grpo}}
+ \lambda_{\mathrm{gapd}} A_{i,k}^{\mathrm{gapd}}.
\end{equation}
The policy is updated with the GRPO objective over generated response tokens, using the standard PPO-style clipped surrogate~\cite{schulman2017proximal,shao2024deepseekmath}:
\begin{equation}
\begin{aligned}
\mathcal{L}_{\mathrm{fuse}}
&= -\frac{1}{Z}\sum_{i,k} m_{i,k}
\min\Bigl(
r_{i,k}(\theta)\widehat{A}_{i,k},
\\
&\qquad\qquad
\mathrm{clip}(r_{i,k}(\theta),1-\epsilon,1+\epsilon)
\widehat{A}_{i,k}
\Bigr),
\\
r_{i,k}(\theta)
&= \frac{\pi_{\theta}(y_{i,k}\mid x_{i,k})}
{\pi_{\mathrm{old}}(y_{i,k}\mid x_{i,k})}.
\end{aligned}
\end{equation}
where $m_{i,k}$ is the response-token mask and $Z=\sum_{i,k}m_{i,k}$.
Thus the final-answer reward supplies the shared GRPO signal, while \method adds local guidance only on aligned student-generated tokens.
The numeric example in Figure~\ref{fig:gapd_framework} illustrates this fusion: a wrong relation segment receives both the coarse negative GRPO advantage and a dense negative GAPD advantage on the divergent relation tokens, yielding a stronger update on the actual error span.

\section{Experiments}
\label{sec:experiments}

\begin{table*}[!t]
\caption{Performance on GrailQA. Bold numbers indicate the best result, and underlined numbers indicate the second-best result.}
\label{tab:grailqa_kbqar1}
\centering
\fontsize{7.2pt}{7.8pt}\selectfont
\setlength{\tabcolsep}{1.6mm}
\resizebox{\textwidth}{!}{%
\begin{tabular}{llcccccc|cc}
\toprule
\multirow{2}{*}{\textbf{Method}} & \multirow{2}{*}{\textbf{LLM}} &
\multicolumn{2}{c}{\textbf{I.I.D.}} &
\multicolumn{2}{c}{\textbf{Compositional}} &
\multicolumn{2}{c|}{\textbf{Zero-shot}} &
\multicolumn{2}{c}{\textbf{Overall}} \\
\cmidrule(lr){3-4} \cmidrule(lr){5-6} \cmidrule(lr){7-8} \cmidrule(lr){9-10}
& & \textbf{EM} & \textbf{F1} & \textbf{EM} & \textbf{F1} & \textbf{EM} & \textbf{F1} & \textbf{EM} & \textbf{F1} \\
\midrule
\multicolumn{10}{c}{\textit{Prompting Methods}} \\
\midrule
KB-BINDER~\cite{KB-BINDER} & Codex-davinci-002 & 40.0 & 43.3 & 33.9 & 36.6 & 40.1 & 44.0 & 38.7 & 42.2 \\
KB-Coder~\cite{KB-Coder} & GPT-3.5-turbo & 40.6 & 45.5 & 34.5 & 38.6 & 42.2 & 47.3 & 40.1 & 44.9 \\
ARG-KBQA~\cite{ARG-KBQA} & GPT-3.5-turbo & 46.6 & 51.5 & 36.4 & 41.8 & 46.6 & 52.1 & 43.8 & 48.5 \\
\multirow{4}{*}{\kbqarone Harness~\cite{kbqa-r1}} & GLM-5 & 46.0 & 62.1 & 51.4 & 64.4 & 68.7 & 75.5 & 60.0 & 70.2 \\
& Kimi-K2.5 & 64.0 & 70.1 & 57.1 & 65.7 & 73.9 & 78.0 & 68.5 & 73.9 \\
& gpt-5.3-codex & 55.6 & 64.3 & 56.2 & 65.6 & 77.4 & 79.6 & 68.6 & 73.5 \\
& Claude 4.6 Sonnet & 62.0 & 68.6 & 62.9 & 68.8 & 79.1 & 81.9 & 72.0 & 76.3 \\
\midrule
\multicolumn{10}{c}{\textit{Fine-tune-based Methods}} \\
\midrule
RnG-KBQA~\cite{RnG-KBQA} & T5-large & 86.7 & 89.0 & 61.7 & 68.9 & 68.8 & 74.7 & 69.5 & 76.9 \\
DecAF~\cite{DecAF} & T5-large & 88.7 & \underline{92.4} & 71.5 & 79.8 & 65.9 & 77.3 & 72.5 & 81.4 \\
TIARA~\cite{TIARA} & T5-large & 88.4 & 91.2 & 66.4 & 74.8 & 73.3 & 80.7 & 75.3 & 81.9 \\
KBQA-o1~\cite{luo2025kbqa} & Llama-3.1-8B & 77.8\smallpm{0.5} & 85.5\smallpm{0.4} & 76.3\smallpm{0.6} & 77.6\smallpm{0.5} & 68.1\smallpm{0.8} & 76.1\smallpm{0.4} & 71.9\smallpm{0.3} & 78.5\smallpm{1.0} \\
KnowCoder-A1~\cite{knowcoder-a1} & Qwen2.5-Coder-7B & -- & 81.1 & -- & 70.6 & -- & 84.1 & -- & 80.5 \\
\kbqarone~\cite{kbqa-r1} & Llama-3.1-8B & \underline{90.0}\smallpm{0.3} & 91.5\smallpm{0.2} & \underline{78.0}\smallpm{0.4} & \underline{82.5}\smallpm{0.3} & \underline{83.6}\smallpm{0.3} & \underline{85.2}\smallpm{0.3} & \underline{83.9}\smallpm{0.2} & \underline{86.1}\smallpm{0.3} \\
\midrule
\rowcolor{gapdrow}
\method & Llama-3.1-8B & \textbf{92.8}\smallpm{0.2} & \textbf{93.8}\smallpm{0.2} & \textbf{82.4}\smallpm{0.3} & \textbf{87.0}\smallpm{0.3} & \textbf{87.8}\smallpm{0.2} & \textbf{88.7}\smallpm{0.2} & \textbf{87.8}\smallpm{0.2} & \textbf{89.5}\smallpm{0.2} \\
\rowcolor{gray!15}
\textit{Rel. improv. over KBQA-R1} & & \textit{+3.1\%} & \textit{+2.5\%} & \textit{+5.6\%} & \textit{+5.5\%} & \textit{+5.0\%} & \textit{+4.1\%} & \textit{+4.6\%} & \textit{+3.9\%} \\
\bottomrule
\end{tabular}
}
\end{table*}

\begin{table*}[!t]
\centering
\begin{minipage}[t]{0.48\textwidth}
\centering
\caption{Results on WebQSP. Bold numbers indicate the best result, and underlined numbers indicate the second-best result.}
\label{tab:webqsp_kbqar1}
\fontsize{8pt}{8.5pt}\selectfont
\setlength{\tabcolsep}{1.6mm}
\resizebox{\linewidth}{!}{%
\begin{tabular}{llc}
\toprule
\textbf{Method} & \textbf{LLM} & \textbf{F1} \\
\midrule
\multicolumn{3}{c}{\textit{Prompting Methods}} \\
\midrule
KB-BINDER~\cite{KB-BINDER} & Codex-davinci-002 & 52.6 \\
KB-Coder~\cite{KB-Coder} & GPT-3.5-turbo & 55.7 \\
ARG-KBQA~\cite{ARG-KBQA} & GPT-3.5-turbo & 58.8 \\
Interactive-KBQA~\cite{Interactive-KBQA} & GPT-4-turbo & 71.2 \\
\multirow{4}{*}{\kbqarone Harness~\cite{kbqa-r1}} & GLM-5 & 64.5 \\
& Kimi-K2.5 & 73.3 \\
& gpt-5.3-codex & 71.1 \\
& Claude 4.6 Sonnet & 76.7 \\
\midrule
\multicolumn{3}{c}{\textit{Fine-tune-based Methods}} \\
\midrule
RnG-KBQA~\cite{RnG-KBQA} & T5-large & 75.6 \\
DecAF~\cite{DecAF} & T5-large & 76.7 \\
TIARA~\cite{TIARA} & T5-large & 78.9 \\
MCTS-KBQA~\cite{MCTS-KBQA} & Llama-3.1-8B & 76.0 \\
KBQA-o1~\cite{luo2025kbqa} & Llama-3.1-8B & 57.8 \\
KnowCoder-A1~\cite{knowcoder-a1} & Qwen2.5-Coder-7B & 77.2 \\
\kbqarone~\cite{kbqa-r1} & Llama-3.1-8B & \underline{83.4}\smallpm{0.3} \\
\midrule
\rowcolor{gapdrow}
\method & Llama-3.1-8B & \textbf{86.1}\smallpm{0.2} \\
\rowcolor{gray!15}
\textit{Rel. improv. over KBQA-R1} & & \textit{+3.2\%} \\
\bottomrule
\end{tabular}
}
\end{minipage}
\hfill
\begin{minipage}[t]{0.48\textwidth}
\centering

\caption{Results on GraphQ. Bold numbers indicate the best result, and underlined numbers indicate the second-best result.}
\label{tab:graphq_kbqar1}
\fontsize{8pt}{8.5pt}\selectfont
\setlength{\tabcolsep}{1.6mm}
\resizebox{\linewidth}{!}{%
\begin{tabular}{llc}
\toprule
\textbf{Method} & \textbf{LLM} & \textbf{F1} \\
\midrule
\multicolumn{3}{c}{\textit{Prompting Methods}} \\
\midrule
KB-BINDER~\cite{KB-BINDER} & Codex-davinci-002 & 27.1 \\
KB-Coder~\cite{KB-Coder} & GPT-3.5-turbo & 31.1 \\
\multirow{4}{*}{\kbqarone Harness~\cite{kbqa-r1}} & GLM-5 & 48.5 \\
& Kimi-K2.5 & 53.3 \\
& gpt-5.3-codex & 51.5 \\
& Claude 4.6 Sonnet & 51.8 \\
\midrule
\multicolumn{3}{c}{\textit{Fine-tune-based Methods}} \\
\midrule
SPARQA~\cite{SPARQA} & BERT-base & 21.5 \\
BERT+Ranking~\cite{GrailQA} & BERT-base & 25.0 \\
ArcaneQA~\cite{ArcaneQA} & BERT-base & 31.8 \\
CoTKR~\cite{CoTKR} & Llama-3-8B & 47.5 \\
KBQA-o1~\cite{luo2025kbqa} & Llama-3.1-8B & 48.7 \\
\kbqarone~\cite{kbqa-r1} & Llama-3.1-8B & \underline{53.8}\smallpm{0.7} \\
\midrule
\rowcolor{gapdrow}
\method & Llama-3.1-8B & \textbf{55.1}\smallpm{0.4} \\
\rowcolor{gray!15}
\textit{Rel. improv. over KBQA-R1} & & \textit{+2.4\%} \\
\bottomrule
\end{tabular}
}
\end{minipage}
\end{table*}

\subsection{Experimental Setup}

\subsubsection{Datasets}

We evaluate on three widely used KBQA benchmarks.
\textbf{GrailQA}~\cite{GrailQA} is designed to evaluate compositional generalization in KBQA.
It contains 64,331 questions, with 44,337 training questions, 6,763 validation questions, and 13,231 test questions.
Following prior work~\cite{ChatKBQA,luo2025kbqa}, we report results on the dev set and break down performance into \textit{i.i.d.}, \textit{compositional}, and \textit{zero-shot} splits.
\textbf{WebQSP}~\cite{WebQSP} contains 4,737 questions with semantic parses, split into 3,098 training questions and 1,639 test questions.
\textbf{GraphQ}~\cite{GraphQ} contains 5,166 graph-structured questions, with 2,508 training questions and 2,658 test questions.
\subsubsection{Baselines}

We organize baselines into prompting and fine-tuning based KBQA methods.
\textbf{Prompting methods} include KB-BINDER~\cite{KB-BINDER}, KB-Coder~\cite{KB-Coder}, ARG-KBQA~\cite{ARG-KBQA}, and Interactive-KBQA~\cite{Interactive-KBQA}.
These methods rely on in-context learning or tool interaction with strong LLMs.
We further include a \kbqarone Harness comparison, where GLM-5~\cite{glm5}, Kimi-K2.5~\cite{kimi25}, gpt-5.3-codex~\cite{gpt53codex}, and Claude 4.6 Sonnet~\cite{claude46sonnet} are plugged into the \kbqarone interaction harness~\cite{kbqa-r1}.
To control the token cost of evaluating recent proprietary models in multi-turn KB interaction, we randomly sample 500 queries for this comparison: 200 GrailQA, 200 WebQSP, and 100 GraphQ examples.
We report these results as \kbqarone Harness rows to make the comparison with current commercial LLMs explicit.
\textbf{Fine-tuning based methods} include RnG-KBQA~\cite{RnG-KBQA}, DecAF~\cite{DecAF}, TIARA~\cite{TIARA}, MCTS-KBQA~\cite{MCTS-KBQA}, KBQA-o1~\cite{luo2025kbqa}, and KnowCoder-A1~\cite{knowcoder-a1}.
For GraphQ, we additionally include SPARQA~\cite{SPARQA}, BERT+Ranking~\cite{GrailQA}, ArcaneQA~\cite{ArcaneQA}, and CoTKR~\cite{CoTKR}.
Detailed baseline descriptions are provided in Appendix~\ref{sec:appendix_baseline_details}.

\subsubsection{Evaluation Metrics}

We use standard KBQA metrics.
\textbf{Exact Match (EM)} measures whether the predicted answer set exactly matches the gold answer set.
\textbf{F1} computes entity-level precision and recall over executed answers.
Training configuration details are provided in Appendix~\ref{sec:appendix_training_config}.

\subsection{Main Results}

Tables~\ref{tab:grailqa_kbqar1}--\ref{tab:graphq_kbqar1} show that \method achieves the best result on every reported benchmark metric.
On GrailQA, it raises overall EM/F1 from 83.9/86.1 to 87.8/89.5, corresponding to relative gains of 4.6\%/3.9\%, and improves every split; the gains are especially clear on compositional and zero-shot questions, where local errors in intermediate relations often determine the final answer.
On WebQSP, \method improves F1 from 83.4 to 86.1 (+3.2\% relative), and on GraphQ from 53.8 to 55.1 (+2.4\% relative).
The margins are not uniform across datasets, but the direction is consistent: adding token-level guidance during RL improves executed-answer accuracy beyond the outcome-reward baseline.
Compared with the strongest commercial LLMs plugged into the \kbqarone interaction harness, \method is substantially stronger on WebQSP, GrailQA, and GraphQ, despite using an 8B open backbone.

\subsection{Ablations and Training Analysis}

\begin{figure*}[t]
    \centering
    \begin{minipage}[t]{0.62\textwidth}
        \centering
        \includegraphics[width=\linewidth]{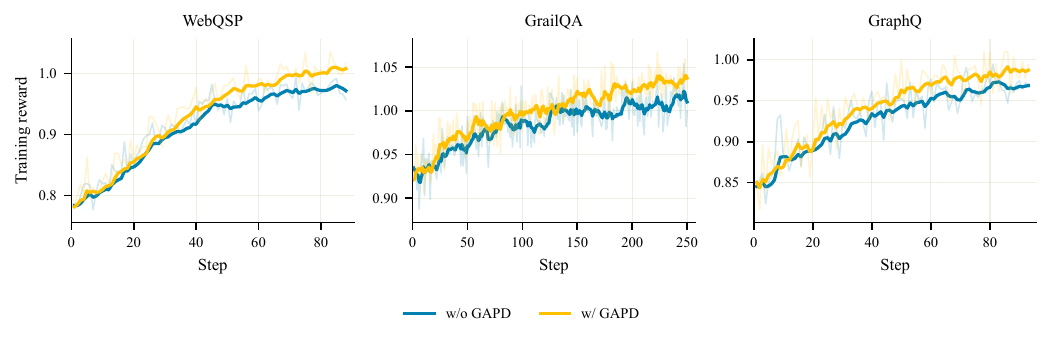}
        \vspace{-1mm}
        {\footnotesize (a) Training reward curves}
    \end{minipage}
    \hfill
    \begin{minipage}[t]{0.34\textwidth}
        \centering
        \includegraphics[width=\linewidth]{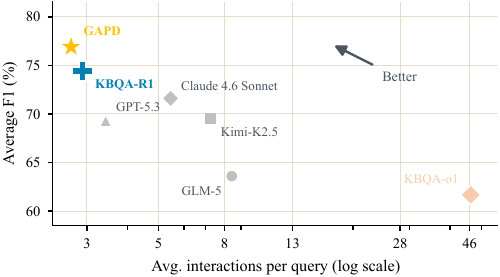}
        \vspace{-1mm}
        {\footnotesize (b) Interaction efficiency}
    \end{minipage}
    \caption{Training dynamics and interaction efficiency.
    Left: reward curves comparing w/o GAPD and w/ GAPD runs; each curve plots mean reward (outcome reward + format reward), with faint per-step values and EMA-smoothed values using $\alpha=0.2$.
    Right: interaction-efficiency comparison, reporting average F1 over three benchmarks against average interactions or calls per query on a log scale.}
    \label{fig:gapd_reward_curves}
    \label{fig:rebuttal_turn_accuracy_tradeoff}
\end{figure*}

We next isolate which components support the gains in Tables~\ref{tab:grailqa_kbqar1}--\ref{tab:graphq_kbqar1}.
Table~\ref{tab:planned_ablation} compares full \method with outcome-only RL, teacher variants, and alignment ablations.
Figure~\ref{fig:gapd_reward_curves} provides optimization and interaction-efficiency analysis.

\begin{table}[t]
\centering
\caption{Ablation study for \method.}
\label{tab:planned_ablation}
\fontsize{8pt}{8pt}\selectfont
\setlength{\tabcolsep}{1.8mm}
\resizebox{\linewidth}{!}{%
\begin{tabular}{l|ccc}
\toprule
\textbf{Variant} & \textbf{WebQSP} & \textbf{GrailQA} & \textbf{GraphQ} \\
\midrule
\rowcolor{gapdrow}
\method & \textbf{86.1} & \textbf{89.5} & \textbf{55.1} \\
~~Outcome reward only & 83.4 & 86.1 & 53.8 \\
\midrule
\multicolumn{4}{l}{\textit{Teacher Variants~\cite{qwen3}}} \\
\midrule
~~Qwen3-32B & 84.8 & 87.2 & 54.1 \\
~~Qwen3-14B & 83.9 & 86.2 & 52.9 \\
\midrule
\multicolumn{4}{l}{\textit{Alignment Ablations}} \\
\midrule
~~\shortstack[l]{Turn-index alignment\\(w/o \midanchor)} & 79.3 & 81.2 & 49.2 \\
~~First-turn only & 85.0 & 87.9 & 54.7 \\
\bottomrule
\end{tabular}}
\end{table}

Table~\ref{tab:planned_ablation} shows three ablation patterns.
First, outcome reward alone is consistently lower, showing that token-level \method supervision provides useful optimization signal beyond the rollout-level reward.
Second, Qwen3-32B and Qwen3-14B teacher variants~\cite{qwen3}, both trained with the SFT dataset, improve over outcome-only RL on most datasets but remain below full \method.
This suggests that policy-consistent gold-action-conditioned self-teacher distributions are more useful than simply using a larger external teacher.
Because the self-teacher reuses the current policy under gold-action conditioning, it also avoids loading an additional teacher model and requires much less GPU memory than the Qwen3 teacher variants.
Third, alignment is critical: turn-index alignment without \midanchor sharply reduces F1, while first-turn-only supervision recovers much of the score but still trails full \method.
The latter result is expected because first turns are naturally aligned, but later turns need state-based matching to assign the token-level guide to the correct branch.

Figure~\ref{fig:gapd_reward_curves}(a) supports the same interpretation from the optimization side.
The reward curves show that adding \method leads to higher training rewards while preserving smooth on-policy optimization.

Figure~\ref{fig:rebuttal_turn_accuracy_tradeoff}(b) evaluates each method as a complete KBQA system by plotting average F1 against the average number of interactions or calls per query.
\method occupies the upper-left region with 76.9 average F1 and 2.68 interactions per query.
This is more accurate and slightly cheaper than the \kbqarone baseline (74.4 F1, 2.91 interactions), and it is also substantially stronger than the best recent LLM harness run, Claude 4.6 Sonnet (71.6 F1, 5.45 interactions).
Thus the gain is not obtained by spending more environment calls, but by making the learned policy more accurate within a compact action trajectory.

Coefficient sensitivity is reported in Appendix~\ref{sec:appendix_lambda_sensitivity}; performance is stable across $\lambda_{\mathrm{gapd}}=0.1$--$1.0$.

\section{Case Study}
\label{sec:case-study}

\begin{figure*}[t]
\centering
\includegraphics[width=\textwidth]{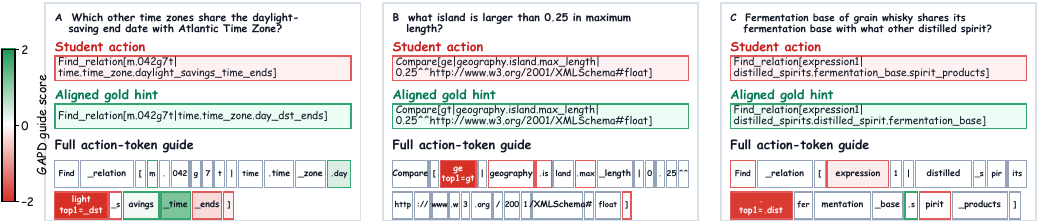}
\caption{Case studies of GAPD.
Each example shows the student action aligned to a gold action and highlights
where token-level guidance is applied on the divergent span.}
\label{fig:gapd_compact_case_studies}
\end{figure*}

Figure~\ref{fig:gapd_compact_case_studies} compresses three dashboard examples
into a paper-facing view; full traces and token tables are in
Appendix~\ref{app:full_case_traces}.
Each example shows the student action, the aligned gold action, and
the token-level guide on the matched student-generated span.

The three cases illustrate complementary failure modes.
In the time-zone example, the student samples a non-canonical daylight-saving
relation surface that is repaired by relation retrieval, and the guide
concentrates on the divergent relation tokens rather than uniformly across the whole generated turn.
In the island comparison example, the question asks for a strict
\emph{larger-than} constraint, while the student uses a non-strict comparator;
the strongest negative score lands exactly on the comparator token.
In the grain-whisky example, \midanchor first matches the intermediate entity
state before applying a gold action, so the penalty is assigned to the
wrong branch rather than to the already-correct first action.

\section{Related Work}

\paragraph{Knowledge Base Question Answering.}
Classic KBQA systems map questions to executable logical forms such as SPARQL or S-expressions and execute them over a KB~\cite{SP,GrailQA}.
LLM prompting methods use demonstrations or code-style representations to generate and bind logical forms~\cite{KB-BINDER,KB-Coder}.
Fine-tuned KBQA generators combine schema retrieval, ranking, or auxiliary grounding objectives with supervised logical-form generation~\cite{TIARA,RnG-KBQA,ChatKBQA}.
Interactive and agentic KBQA systems expose KB operations as tool calls, search actions, or RL rollouts~\cite{Interactive-KBQA,luo2025kbqa,kbqa-r1,knowcoder-a1}.
Subgraph-based methods retrieve a question-specific KB neighborhood before reasoning: earlier systems use graph neural reasoning or iterative retrieval~\cite{GRAFT-Net,PullNet,NSM}, while recent retrievers constrain the graph context passed to the reasoner or LLM~\cite{SR,SubgraphRAG}.
Our work focuses on the RL training stage of action-based KBQA and asks how gold logical forms can provide dense supervision during on-policy updates.

\paragraph{RL for LLM Reasoning.}
RL methods such as PPO~\cite{schulman2017proximal} and \grpo~\cite{shao2024deepseekmath} optimize LLM policies using scalar rewards.
In reasoning and tool-use tasks, these rewards are often sparse and outcome-based.
Recent agentic KBQA work also explores outcome-supervised training without process-level trajectory supervision~\cite{knowcoder-a1}.
We complement outcome-based RL with token-level supervision derived from executable gold actions.


\paragraph{On-Policy Distillation.}
Knowledge distillation transfers soft teacher distributions to a student model~\cite{hinton2015distilling}.
Classical distillation usually trains on fixed teacher-generated or dataset prefixes, whereas on-policy distillation samples from the student's current policy and obtains feedback on the trajectories the student actually visits~\cite{song2026surveyopd}.
This idea is closely related to interactive imitation learning, where querying an expert on learner-induced states reduces distribution shift between training and deployment~\cite{ross2011dagger}.
For LLMs, GKD trains on student-generated outputs with teacher feedback and supports several divergence choices~\cite{agarwal2024gkd}, MiniLLM optimizes reverse KL through an on-policy objective~\cite{gu2024minillm}.
\method follows the same on-policy premise but targets RL-based KBQA: GRPO samples the student rollout, executable gold logical forms are aligned to the student's execution state, and the added distillation signal is applied only to matched turns.

\section{Conclusion}

We proposed \method, a framework for using gold KBQA actions during on-policy RL.
Instead of relying only on final outcome reward, the method conditions the current model on aligned gold actions to produce self-teacher distributions and distills token-level guidance back into the student policy.
This provides a path toward dense and stable RL training for agentic KBQA.

\section*{Limitations}

One practical limitation is training cost.
Compared with outcome-reward-only RL, GAPD training is roughly 1.5$\times$ slower because the gold-action-conditioned self-teacher must score generated student tokens on supervised spans; Appendix Table~\ref{tab:appendix_train_test_tradeoff} summarizes this train-time and test-time tradeoff.
This overhead is a training-time tradeoff for finer token-level supervision.
It does not introduce an external teacher model or any additional computation at inference time.
Instead, the learned policy navigates the knowledge graph more precisely and uses fewer LLM calls per query on average in our experiments, which can make testing faster despite the higher training cost.

\bibliography{refs}

\appendix
\section{Training Configuration}
\label{sec:appendix_training_config}

\textbf{Backbone.}
All reported \method runs use Llama-3.1-8B-Instruct as the policy backbone.

\textbf{SFT warm-start and RL.}
Training uses a two-stage recipe.
First, the policy is warm-started by SFT on action trajectories generated from logical-form annotations, including referenced rejection sampling (\rrs).
Second, the policy is optimized with executable KB feedback using \grpo and token-level \method guidance on aligned student-generated token spans.
The SFT stage uses a learning rate of $5\times10^{-6}$ with cosine decay.
The \grpo stage uses an actor learning rate of $1\times10^{-6}$ and 30\% linear warmup.
GrailQA is trained for 1 epoch due to its larger training set, while WebQSP and GraphQ are trained for 8 epochs.
The train batch size is 256, and we report the final training-step checkpoint.

\textbf{GRPO configuration.}
The rollout policy samples $n=5$ responses per prompt with temperature 1.0 and top-$p=0.99$.
The PPO-style clipping ratios are $\epsilon_{\mathrm{low}}=0.2$ and $\epsilon_{\mathrm{high}}=0.28$, following DAPO~\cite{yu2025dapo}.
The KL coefficient is $\beta=0.001$.
The reward weights are $\lambda_{\mathrm{outcome}}=1.0$, $\lambda_{\mathrm{format}}=0.1$, and $\lambda_{\mathrm{gapd}}=0.5$; raw \method guide scores are clipped to $[-2,2]$ before scaling.
For \midanchor, we accept a matched gold state only when the Jaccard state-match score is at least $\tau_{\mathrm{state}}=0.95$.
The PPO mini-batch size is 128 with dynamic micro-batching enabled.

\textbf{Infrastructure.}
Training is conducted on 8 NVIDIA A100-80GB GPUs with FSDP model sharding.
The Freebase backend uses Virtuoso with ODBC connection pooling, pool size 48, and query timeout 600 seconds.

\begin{table}[t]
    \centering
    \footnotesize
    \setlength{\tabcolsep}{1.6mm}
    \begin{tabular}{@{}lccc@{}}
        \toprule
        \textbf{Method} & \textbf{Train time} & \textbf{Avg. F1} & \textbf{Avg. test calls} \\
        \midrule
        Outcome-only RL & 8.2h & 74.4 & 2.91 \\
        \method & 12.4h & \textbf{76.9} & \textbf{2.68} \\
        \bottomrule
    \end{tabular}
    \caption{Training-time and test-efficiency tradeoff.
    GAPD incurs additional training cost because the gold-action-conditioned self-teacher scores student tokens on supervised spans.
    At test time, no teacher or gold action is used; the trained policy requires fewer LLM calls per query on average while achieving higher average F1.
    Training time is measured on WebQSP for 88 RL steps; average F1 and test calls are computed over WebQSP, GrailQA, and GraphQ.}
    \label{tab:appendix_train_test_tradeoff}
\end{table}

\section{Training Algorithm}
\label{sec:appendix_training_algorithm}

Algorithm~\ref{alg:gapd} summarizes one on-policy training step of \method.
The procedure starts from an ordinary student rollout under the current policy,
executes both the sampled student actions and the gold action sequence, and then
uses \midanchor to identify student turns whose reached entity states can be
aligned to gold states.
Only these aligned turns receive gold-action-conditioned self-teacher scores.

\begin{algorithm}[t]
\small
\caption{\method Training Step}
\label{alg:gapd}
\begin{algorithmic}[1]
\Require prompt $q$, gold actions $\mathbf{g}$, policy $\pi_\theta$, KB executor $\mathcal{E}$
\State Roll out the student policy to obtain trajectory $\tau$
\State Execute student actions and collect entity-anchor states $\{E(s_t)\}$
\State Execute gold actions and collect entity-anchor states $\{E(u_m)\}$
\State Initialize distillation mask $\mathcal{I}\leftarrow\emptyset$
\For{each student turn $t$}
  \State Match $s_t$ to gold state $u_{m^*(t)}$ using reached entity sets
  \State Build teacher-side prompt $\tilde{x}_t$ by adding gold action $g_{m^*(t)+1}$
  \State Add student-generated token positions at turn $t$ to $\mathcal{I}$
\EndFor
\State Compute GRPO advantages from final answer reward
\State Score student tokens under the stop-gradient self-teacher
\State Convert teacher-student log-prob gaps into token-level \method advantages on $\mathcal{I}$
\State Update $\theta$ with a GRPO-style clipped loss using fused token advantages
\end{algorithmic}
\end{algorithm}

\section{Coefficient Sensitivity}
\label{sec:appendix_lambda_sensitivity}

\begin{figure}[t]
    \centering
    \includegraphics[width=\linewidth]{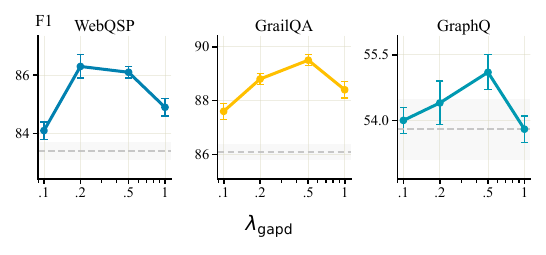}
    \caption{Sensitivity to the \method coefficient $\lambda_{\mathrm{gapd}}$.
    Each panel reports F1 after varying $\lambda_{\mathrm{gapd}}$; the dashed gray horizontal line marks $\lambda_{\mathrm{gapd}}=0$, i.e., the \kbqarone outcome-only RL baseline without \method.}
    \label{fig:gapd_lambda_sensitivity}
\end{figure}

The coefficient sweep shows a useful range around $\lambda_{\mathrm{gapd}}=0.1$--$1.0$: very small weights underuse the guide, while the gains taper at $\lambda_{\mathrm{gapd}}=1.0$ compared with the stronger $\lambda_{\mathrm{gapd}}=0.5$ setting.
Overall, performance remains stable across a broad coefficient range, indicating that \method is not overly sensitive to this hyperparameter.

\begin{figure}[h]
    \centering
    \includegraphics[width=\linewidth]{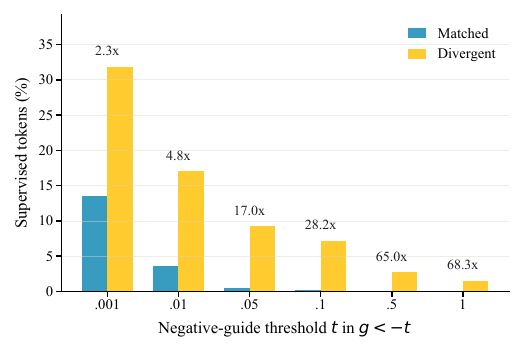}
    \caption{Negative-tail mass of the token-level \method guide.
    Bars show the share of supervised generated tokens whose raw guide score $g$ falls below $-t$.}    \label{fig:gapd_tail_mass}
\end{figure}

Figure~\ref{fig:gapd_tail_mass} provides an additional guide-strength analysis.
It reports the share of supervised generated tokens with strong negative raw guide scores.
Divergent actions consistently have heavier negative tails, indicating that the token-level guide mainly acts as a localized suppressive correction on generated spans that depart from the gold-action-conditioned self-teacher.

\FloatBarrier

\section{Baseline Details}
\label{sec:appendix_baseline_details}

We compare with representative KBQA systems under the settings reported by their original papers or by the corresponding interaction harness.
The baselines cover prompting-only systems, fine-tuned semantic parsers, and recent action-based or outcome-supervised KBQA agents.

\begin{table*}[t]
    \centering
    \small
    \resizebox{\linewidth}{!}{%
    \begin{tabular}{llll}
        \toprule
        \textbf{Action} & \textbf{Arguments} & \textbf{Target Function} & \textbf{Equivalent Logical Form} \\
        \midrule
        \texttt{Find\_relation} & \textit{entity} $\vert$ \textit{relation} & \textit{expression} = JOIN(`\textit{relation}', START(\textit{entity})) & (JOIN \textit{relation} \textit{entity}) \\
        \texttt{Merge} & \textit{expression1} $\vert$ \textit{expression} & \textit{expression} = AND(\textit{expression1}, \textit{expression}) & (AND (\textit{expression1}) (\textit{expression})) \\
        \texttt{Order} & \textit{MAX/MIN} $\vert$ \textit{expression} $\vert$ \textit{relation} & \textit{expression} = ARG(`\textit{mode}', \textit{expression}, `\textit{relation}') & (\textit{mode} (\textit{expression}) \textit{relation}) \\
        \texttt{Compare} & \textit{le/lt/ge/gt} $\vert$ \textit{relation} $\vert$ \textit{number} & \textit{expression} = CMP(`\textit{mode}', `\textit{relation}', \textit{number}, \textit{expression}) & (\textit{mode} \textit{relation} \textit{number} (\textit{expression})) \\
        \texttt{Time\_constraint} & \textit{relation} $\vert$ \textit{time} & \textit{expression} = TC(\textit{expression}, `\textit{relation}', `\textit{time}') & (TC (\textit{expression}) \textit{relation} \textit{time}) \\
        \texttt{Count} & \textit{expression} & \textit{expression} = COUNT(\textit{expression}) & (COUNT (\textit{expression})) \\
        \bottomrule
    \end{tabular}%
    }
    \caption{Action space used in the action-based KBQA setting. Each action is mapped to a functional update over the intermediate expression state and an equivalent logical-form template; the resulting S-expression is later translated into SPARQL for execution.}
    \label{tab:appendix_action_space}
\end{table*}

\paragraph{Prompting methods.}
These systems use strong LLMs with in-context examples, retrieved KB context, or tool interaction, but do not train a task-specific policy with executable outcome reward.
\textbf{KB-BINDER}~\cite{KB-BINDER} uses few-shot logical-form demonstrations to produce a draft program and then binds the draft to executable KB items.
\textbf{KB-Coder}~\cite{KB-Coder} converts logical-form generation into a code-style in-context learning problem to reduce format errors.
\textbf{ARG-KBQA}~\cite{ARG-KBQA} augments LLM prompting with question-related graph structures retrieved from the KB.
\textbf{Interactive-KBQA}~\cite{Interactive-KBQA} exposes KB interaction APIs to the LLM and decomposes complex questions through multi-turn tool use.
\textbf{ToG}~\cite{ToG} and \textbf{PoG}~\cite{PoG} are prompting-based graph reasoning agents that perform test-time graph exploration or planning; their reported Hits@1 results are not mixed into our F1/EM main tables.

\paragraph{\kbqarone Harness with commercial LLMs.}
To test whether recent general-purpose LLMs can solve the task without KBQA-specific RL, we plug GLM-5~\cite{glm5}, Kimi-K2.5~\cite{kimi25}, gpt-5.3-codex~\cite{gpt53codex}, and Claude 4.6 Sonnet~\cite{claude46sonnet} into the \kbqarone interaction harness~\cite{kbqa-r1}.
These runs use the same action-style KB interface and executor, but the underlying LLMs are not fine-tuned on our training data.
Because multi-turn KB execution with proprietary models is expensive, we evaluate this harness comparison on a 500-example subset containing 200 GrailQA, 200 WebQSP, and 100 GraphQ queries.

\paragraph{Fine-tuning and agentic KBQA methods.}
\textbf{RnG-KBQA}~\cite{RnG-KBQA} first ranks candidate logical forms retrieved from KB search and then generates the final logical form conditioned on the ranked candidates.
\textbf{DecAF}~\cite{DecAF} jointly decodes answers and logical forms for KBQA.
\textbf{TIARA}~\cite{TIARA} combines multi-grained retrieval over entities, schema items, and exemplary logical forms with constrained decoding.
\textbf{MCTS-KBQA}~\cite{MCTS-KBQA} uses Monte Carlo tree search with step-wise rewards to explore KBQA reasoning paths with open-source instruction LLMs.
\textbf{KBQA-o1}~\cite{luo2025kbqa} is an agentic KBQA framework that uses Monte-Carlo tree search with policy and reward models to explore KB action trajectories.
\textbf{KnowCoder-A1}~\cite{knowcoder-a1} trains an agentic KBQA policy with outcome supervision and curriculum-style optimization.
\textbf{\kbqarone}~\cite{kbqa-r1} is our primary outcome-reward-only RL baseline: it uses the same Llama-3.1-8B backbone, action space, Freebase executor, and SFT warm-start recipe, but optimizes GRPO with final-answer and format rewards without token-level \method guidance.

\paragraph{GraphQ-specific baselines.}
For GraphQ, we additionally include established semantic-parsing and graph-reasoning baselines reported in prior KBQA work.
\textbf{SPARQA}~\cite{SPARQA} uses skeleton-based semantic parsing for complex KB questions.
\textbf{BERT+Ranking}~\cite{GrailQA} ranks candidate parses with a BERT-based model.
\textbf{ArcaneQA}~\cite{ArcaneQA} combines generation with dynamic program induction and contextual schema encoding.
\textbf{CoTKR}~\cite{CoTKR} uses chain-of-thought knowledge rewriting to convert retrieved KB evidence into reasoning-friendly text.

\section{Action Space}
\label{sec:appendix_action_space}

Following the action-based KBQA setting, we decompose logical-form construction
into a small set of executable operations over an evolving intermediate
expression.
At each turn, the model emits an action inside the structured action block; the
executor parses the action, updates the corresponding functional expression, and
returns KB feedback that is appended to the next-turn context.
This design avoids requiring the policy to generate a complete nested
S-expression in one shot, while still keeping every intermediate step
convertible to an executable logical form.
Table~\ref{tab:appendix_action_space} lists the action templates used in our
rollouts.
The first four columns specify the surface action, its arguments, the functional
state update, and the equivalent logical-form fragment.
In training, \method applies token-level guidance only to matched student-generated spans;
environment feedback remains context rather than a supervised target.

\section{Prompt Templates}
\label{sec:appendix_prompts}

\lstdefinestyle{promptstyle}{
  basicstyle=\ttfamily\scriptsize,
  breaklines=true,
  breakatwhitespace=true,
  columns=fullflexible,
  keepspaces=true,
  frame=single,
  framesep=5pt,
  rulecolor=\color{black!55},
  showstringspaces=false,
  xleftmargin=0pt,
  xrightmargin=0pt,
  aboveskip=0.75\baselineskip,
  belowskip=0.75\baselineskip,
  captionpos=b
}

Tables~\ref{tab:kbqa_prompt_template} and~\ref{tab:teacher_prompt_augmentation} show the two prompt fragments used during training.
The student rollout prompt defines the action protocol, while the teacher-only reference block is inserted only when computing \method token guidance for a matched student state.
At inference time, the teacher-side block is never used.

\begin{table}[H]
    \centering
    \scriptsize
    \begin{tabularx}{\linewidth}{>{\raggedright\arraybackslash}X}
        \toprule\\[-6pt]
        You are an expert assistant for querying the Freebase knowledge base using structured reasoning actions. \\
        Answer the given question about Freebase knowledge base. \\
        You MUST conduct reasoning inside \textcolor{blue}{\texttt{<think>...</think>}} before emitting actions. \\
        After reasoning, provide structured actions inside \textcolor{cyan}{\texttt{<action>...</action>}}. \\
        The system will return query results between \textcolor{magenta}{\texttt{<information>...</information>}}. \\
        When ready, return the final answer inside \textcolor{purple}{\texttt{<answer>...</answer>}} using MIDs or literal values. \\
        \texttt{Available Actions :} \textcolor{red}{\texttt{\{The Description of Actions\}}} \\
        Question: \textcolor{red}{\{QUESTION\}}.\\
        \bottomrule
    \end{tabularx}
    \caption{Action-based reasoning prompt template for the student policy.}
    \label{tab:kbqa_prompt_template}
\end{table}

\begin{table}[H]
    \centering
    \scriptsize
    \begin{tabularx}{\linewidth}{>{\raggedright\arraybackslash}X}
        \toprule\\[-6pt]
        \textit{Student context before the matched generation turn.} \\
        \textcolor{gray!70}{\texttt{\ldots dialogue history, actions, and KB observations \ldots}} \\
        \midrule
        \textbf{Teacher-only reference block} \\
        \textcolor{gray!70}{\texttt{[Reference action sequence. Do not mention or repeat this in your response, just silently use it as guidance.]}} \\
        \textcolor{red}{\texttt{\{matched\_gold\_action\}}} \\
        \textcolor{gray!70}{\texttt{[End reference]}} \\
        \midrule
        \textit{Assistant-turn prefix from the chat template; teacher logits are evaluated on the original student response tokens after this point.} \\
        \bottomrule
    \end{tabularx}
    \caption{Teacher-only prompt augmentation for computing \method token guidance.
    The full dataset-specific action descriptions are inserted at the \texttt{\{The Description of Actions\}} placeholder and summarized in Table~\ref{tab:appendix_action_space}.}
    \label{tab:teacher_prompt_augmentation}
\end{table}

\section{Additional Subgraph-Based Baselines}
\label{sec:appendix_subgraph_baselines}

Table~\ref{tab:appendix_subgraph_webqsp} provides a focused comparison with
subgraph-based KBQA baselines, including RoG, GNN-RAG, and SubgraphRAG.
These methods retrieve or construct question-specific graph neighborhoods and
perform reasoning over the resulting subgraph, so we report them separately as
additional graph-context baselines.
We keep the comparison in the appendix because the available comparable results
are WebQSP-only and therefore do not match the cross-dataset F1/EM setting of
the main tables.
Among the listed subgraph-based baselines, SubgraphRAG with GPT-4o reports the
strongest WebQSP F1 of 78.2; the outcome-only \kbqarone baseline reaches 83.4,
and \method further improves to 86.1.
DoG~\cite{DoG} is discussed in related work but omitted from the table because
we could not identify a directly comparable WebQSP F1 result.

\begin{table}[H]
    \centering
    \small
    \setlength{\tabcolsep}{2.2mm}
    \begin{tabularx}{\linewidth}{@{}>{\raggedright\arraybackslash}X l c@{}}
        \toprule
        \textbf{Method} & \textbf{Backbone} & \textbf{WebQSP F1} \\
        \midrule
        RoG~\cite{RoG} & Llama2-7B & 70.8 \\
        GNN-RAG~\cite{GNN-RAG} & Llama2-7B & 71.3 \\
        SubgraphRAG~\cite{SubgraphRAG} & Llama-3.1-8B & 70.6 \\
        SubgraphRAG~\cite{SubgraphRAG} & GPT-4o & 78.2 \\
        \midrule
        \kbqarone~\cite{kbqa-r1} & Llama-3.1-8B & 83.4 \\
        \rowcolor{gapdrow}
        \method & Llama-3.1-8B & \textbf{86.1} \\
        \bottomrule
    \end{tabularx}
    \caption{Additional WebQSP comparison with graph/subgraph reasoning baselines.}
    \label{tab:appendix_subgraph_webqsp}
\end{table}

\section{Full Student Traces and Teacher Token Scores}
\label{app:full_case_traces}

\definecolor{traceThink}{HTML}{1D4ED8}
\definecolor{traceAction}{HTML}{008B8B}
\definecolor{traceInfo}{HTML}{8B5CF6}
\definecolor{traceAnswer}{HTML}{7E22CE}
\definecolor{scoreNeg}{HTML}{B91C1C}
\definecolor{scorePos}{HTML}{047857}

\newtcolorbox{gapdtracebox}{
  enhanced,
  breakable,
  colback=white,
  colframe=black!40,
  boxrule=0.45pt,
  arc=0pt,
  left=5pt,
  right=5pt,
  top=5pt,
  bottom=5pt,
  fontupper=\scriptsize\raggedright\sloppy,
  before skip=0.5\baselineskip,
  after skip=0.75\baselineskip
}

\newcommand{\traceheader}[1]{\par\smallskip\noindent{\sffamily\bfseries\colorbox{gray!20}{\strut\; #1 \;}}\par\smallskip}
\newcommand{\tracecode}[1]{\par\noindent\texttt{\detokenize{#1}}\par}
\newcommand{\tok}[1]{\texttt{\detokenize{#1}}}
\newcommand{\negscore}[1]{\textcolor{scoreNeg}{#1}}
\newcommand{\posscore}[1]{\textcolor{scorePos}{#1}}
\newcommand{\scorecell}[4]{%
  \begin{minipage}[t]{0.105\textwidth}
  \centering\scriptsize
  \begin{tabular}{@{}c@{}}
  \textcolor{gray!65}{\scriptsize #1}\\[-1pt]
  \tok{#2}\\[-1pt]
  #3\\[-1pt]
  \textcolor{gray!70}{\tok{#4}}
  \end{tabular}
  \end{minipage}%
}
\newcommand{\scorecellraw}[4]{%
  \begin{minipage}[t]{0.105\textwidth}
  \centering\scriptsize
  \begin{tabular}{@{}c@{}}
  \textcolor{gray!65}{\scriptsize #1}\\[-1pt]
  \texttt{#2}\\[-1pt]
  #3\\[-1pt]
  \textcolor{gray!70}{\texttt{#4}}
  \end{tabular}
  \end{minipage}%
}

\begin{figure*}[p]
\centering
\includegraphics[width=\textwidth,height=0.95\textheight,keepaspectratio]{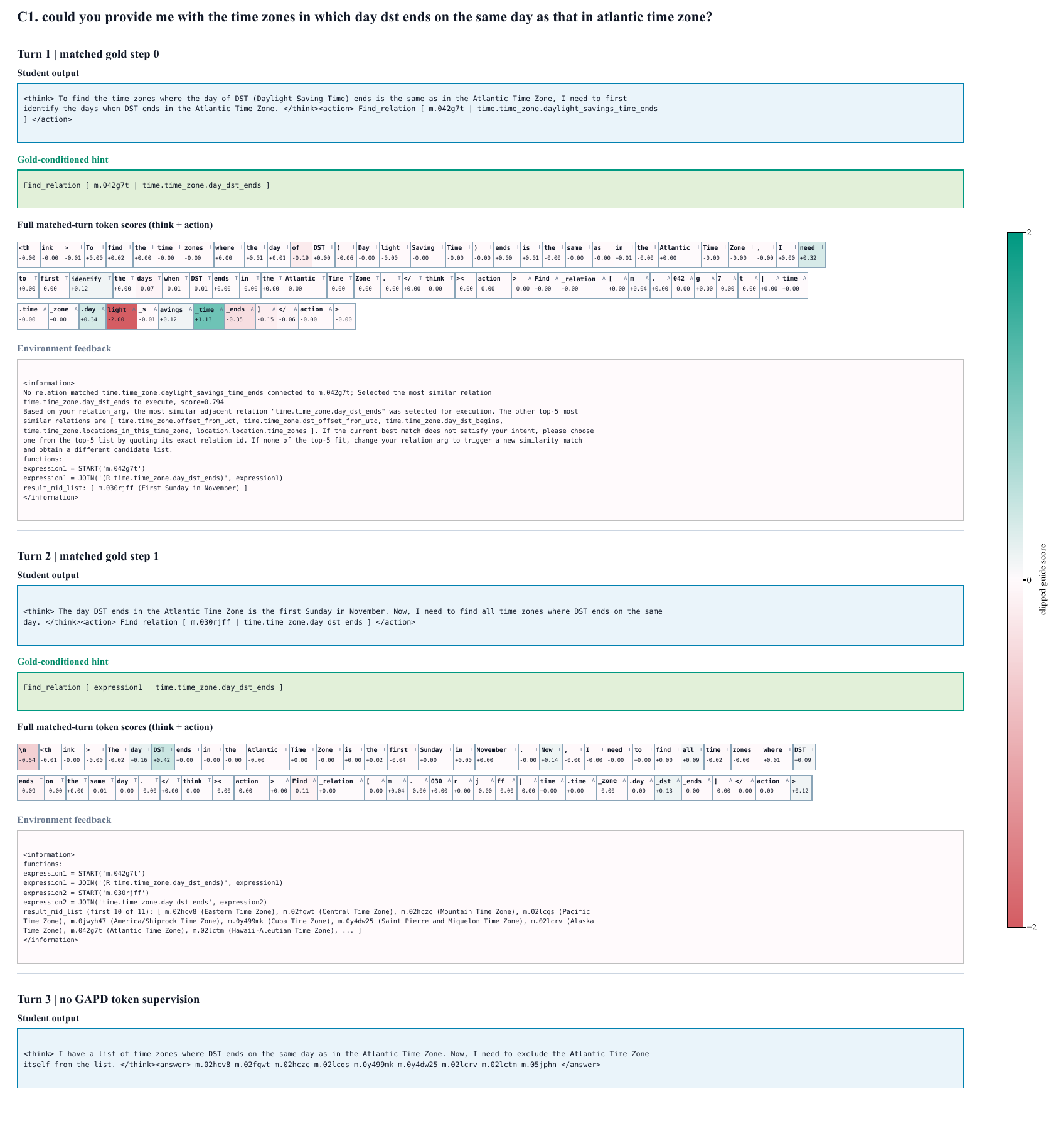}
\caption{Full trajectory view for the time-zone case. The figure preserves the complete student--environment interaction chain and overlays full think-plus-action token scores for supervised turns.}
\label{fig:case_time_zone_full_trajectory}
\end{figure*}

\begin{figure*}[p]
\centering
\includegraphics[width=\textwidth,height=0.95\textheight,keepaspectratio]{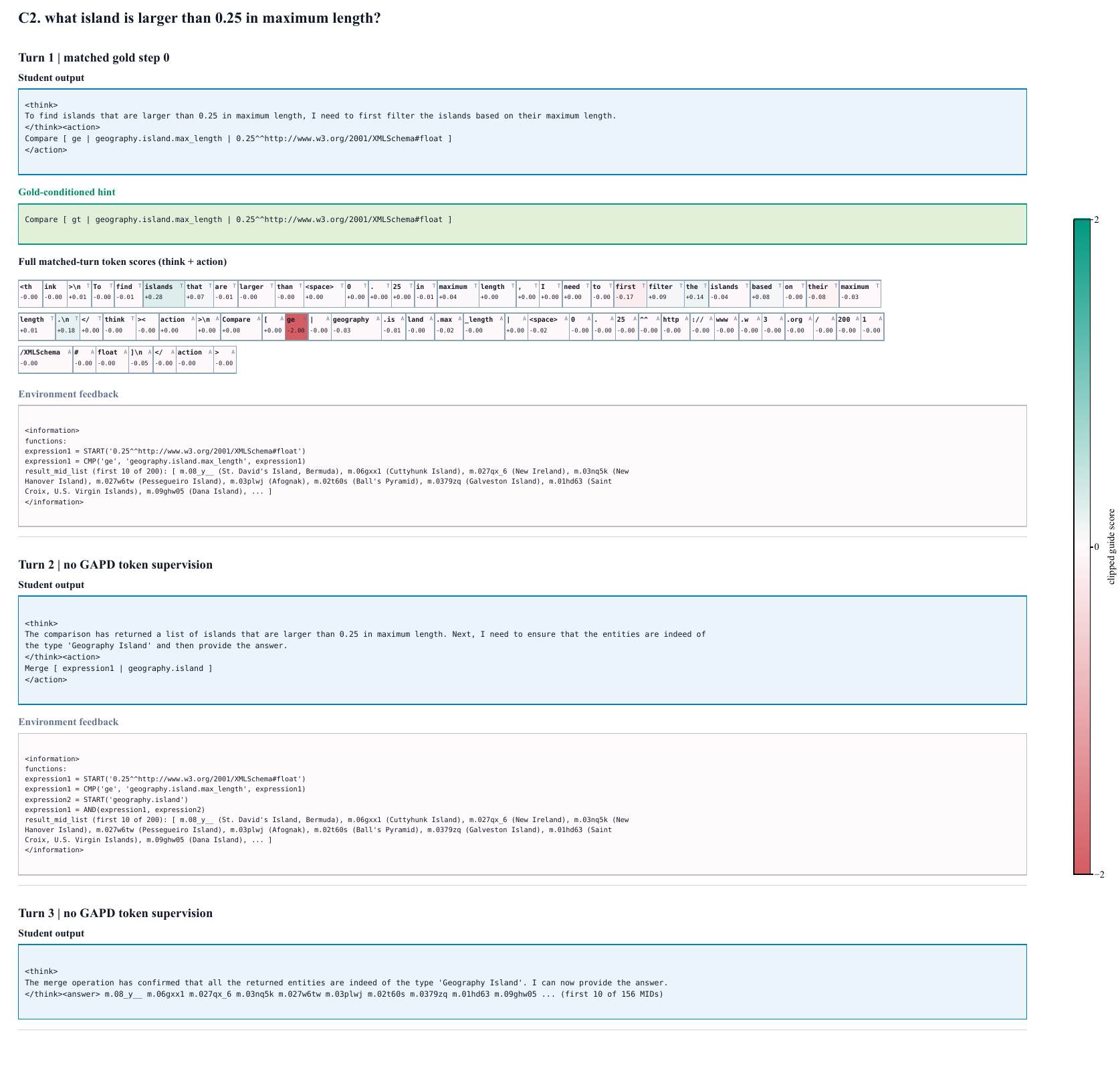}
\caption{Full trajectory view for the island comparison case. The figure preserves the complete student--environment interaction chain and overlays think-plus-action token scores for the supervised comparison turn. For readability, long \texttt{result\_mid\_list} fields and answer MID lists are truncated to the first ten entries, with ellipses marking omitted MIDs.}
\label{fig:case_island_compare_full_trajectory}
\end{figure*}

\begin{figure*}[p]
\centering
\includegraphics[width=\textwidth,height=0.95\textheight,keepaspectratio]{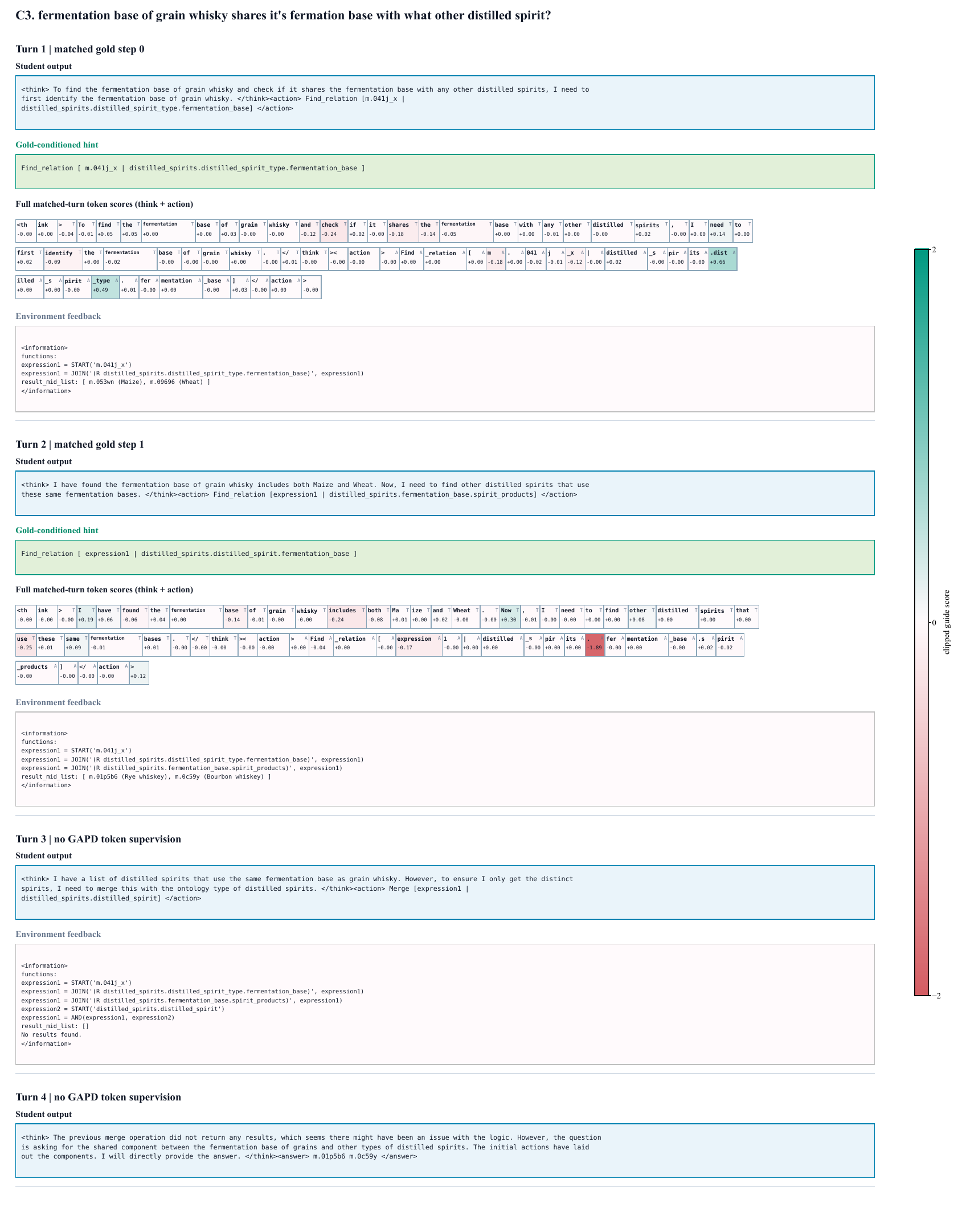}
\caption{Full trajectory view for the distilled-spirits case. The figure preserves the complete student--environment interaction chain and overlays full think-plus-action token scores for supervised turns.}
\label{fig:case_distilled_spirits_full_trajectory}
\end{figure*}

We expand the three dashboard case studies into full appendix trajectory views.
The aligned gold action is visible only to the self-teacher and is used only to compute token-level guide scores for safely matched spans; red and green cells mark negative and positive guide values.
Figures~\ref{fig:case_time_zone_full_trajectory}--\ref{fig:case_distilled_spirits_full_trajectory} provide the full traces for the time-zone relation, strict-comparison, and distilled-spirits cases from the main text.

\phantomsection\label{app:case_time_zone}

\phantomsection\label{app:case_island_compare}

\phantomsection\label{app:case_distilled_spirits}

\section{AI Assistant Use}
We used AI assistants to help modify experimental and analysis code and to polish manuscript wording.
The authors reviewed and verified all generated code, experiments, analyses, and final paper text.

\end{document}